\documentclass[letterpaper, 10 pt, conference]{ieeeconf}
\IEEEoverridecommandlockouts
\usepackage{times}

\usepackage[utf8]{inputenc} 
\usepackage[T1]{fontenc}    
\usepackage{hyperref}       
\usepackage{url}             
\usepackage{amsmath} 
\usepackage{amssymb}  
\usepackage{booktabs}       
\usepackage{amsfonts}       
\usepackage{nicefrac}       
\usepackage{microtype}      
\usepackage{graphicx}
\usepackage{algorithm}
\usepackage{algpseudocode}
\usepackage{subcaption}
\usepackage{multirow}
\usepackage{multicol}
\usepackage{color}

\newcommand{\rebuttal}[1]{#1}

\usepackage{amsmath,amsfonts,bm}

\def\eqref#1{equation~\ref{#1}}

\def\1{\bm{1}}

\def\vmu{{\bm{\mu}}}

\def\ve{{\bm{e}}}

\def\vu{{\bm{u}}}

\def\vx{{\bm{x}}}

\DeclareMathAlphabet{\mathsfit}{\encodingdefault}{\sfdefault}{m}{sl}
\SetMathAlphabet{\mathsfit}{bold}{\encodingdefault}{\sfdefault}{bx}{n}

\DeclareMathOperator*{\argmin}{arg\,min}

\newcommand{\fpe}{{f^{\text{PE}}}}

\newcommand{\fgp}{{f^{\text{GP}}}}
\newcommand{\freal}{{f^{\text{real}}}}

\newcommand{\fpered}{{f^{\text{PE}}_{\text{red}}}}
\newcommand{\fperedmu}{{f^{\text{PE}}_{\text{red},\vmu}}}
\newcommand{\fperedmuopt}{{f^{\text{PE}}_{\text{red},\vmu^\ast}}}
\newcommand{\fpefull}{{f^{\text{PE}}_{\text{full}}}}
\newcommand{\fimm}{{f_{\text{imm}}}}

\title{\LARGE \bf
    \rebuttal{Data-Efficient Learning for Complex and Real-Time Physical Problem Solving using Augmented Simulation}
}

\author{Kei Ota$^{1}$, Devesh K. Jha$^{2}$, Diego Romeres$^{2}$, Jeroen van Baar$^{2}$, Kevin A. Smith$^{3}$, Takayuki Semitsu$^{1}$,\\
Tomoaki Oiki$^{1}$, Alan Sullivan${^2}$, Daniel Nikovski${^2}$, and Joshua B. Tenenbaum$^{3}$\\
$^{1}$Mitsubishi Electric, $^{2}$Mitsubishi Electric Research Labs, $^{3}$Massachusetts Institute of Technology
}

\begin{document}
    \twocolumn[{%
    \renewcommand\twocolumn[1][]{#1}%
    \maketitle
    }]
    \thispagestyle{empty}
    \pagestyle{empty}

    \begin{abstract}
Humans quickly solve tasks in novel systems with complex dynamics, without requiring much interaction. While deep reinforcement learning algorithms have achieved tremendous success in many complex tasks, these algorithms need a large number of samples to learn meaningful policies. In this paper, we present a task for navigating a marble to the center of a circular maze. While this system is very intuitive and easy for humans to solve, it can be very difficult and inefficient for standard reinforcement learning algorithms to learn meaningful policies. We present a model that learns to move a marble in the complex environment within minutes of interacting with the real system. Learning consists of initializing a physics engine with parameters estimated using data from the real system. The error in the physics engine is then corrected using Gaussian process regression, which is used to model the residual between real observations and physics engine simulations. The physics engine augmented with the residual model is then used to control the marble in the maze environment using a model-predictive feedback over a receding horizon. To the best of our knowledge, this is the first time that a hybrid model consisting of a full physics engine along with a statistical function approximator has been used to control a complex physical system in real-time using nonlinear model-predictive control (NMPC). 
\end{abstract}

    \section{Introduction}\label{sec:intro}

\rebuttal{
Artificial Intelligence has long had the goal of designing robotic agents that can interact with the (complex) physical world in flexible, data-efficient and generalizable ways \cite{fang2019learning,toussaint2018differentiable}.
Model-based control methods form plans based on predefined models of the world dynamics. However, although data-efficient, these systems require accurate dynamics models, which may not exist for complex tasks. 
Model-free methods on the other hand rely on reinforcement learning, where the agents simultaneously learn a model of the world dynamics and a control policy~\cite{levine2016,sanchez2018graph}. However, although these methods can learn policies to solve tasks involving complex dynamics, training these policies is inefficient, as they require many samples. Furthermore, these method are typically not generalizable beyond the trained scenarios.
}
\begin{figure}
    \centering
    \includegraphics[width=0.95\columnwidth]{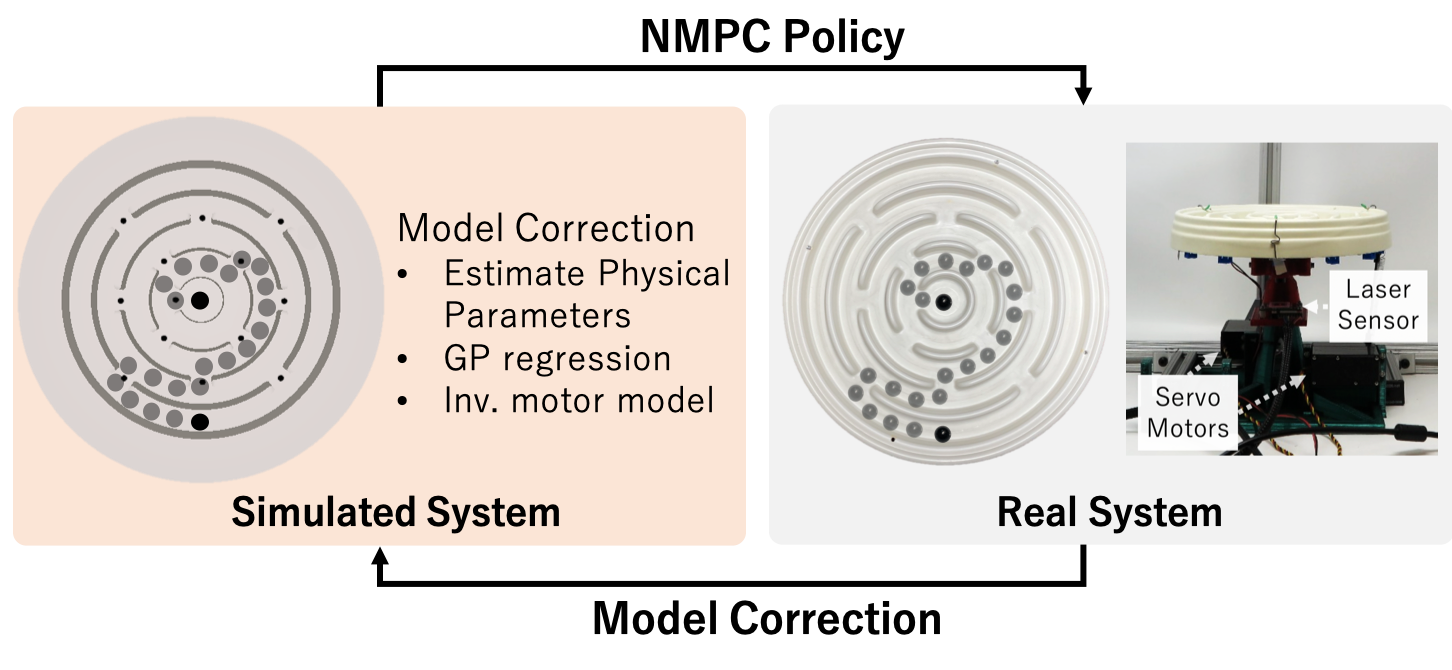}
    \caption{
        \rebuttal{
        We train a reinforcement learning agent that initializes a policy with a general purpose physics engine, then corrects its dynamics model using parameter estimation and residual learning. The agent uses this augmented model in a circular maze to drive a marble to the center.
        }
    }
    \label{fig:learningapproach}
\end{figure}

\rebuttal{
Our aim in this paper is to combine the best of both methodologies: our system uses nonlinear model predictive control with a predefined (inaccurate) model of dynamics at its core, but updates that model by learning residuals between predictions and real-world observations via physical parameter estimation and Gaussian process regression \cite{williams2006gaussian}. 
We take inspiration from cognitive science for this approach, as people can interact with and manipulate novel objects well with little or no prior experience \cite{osiurak2018looking}. Research suggests people have internal models of physics that are well calibrated to the world \cite{battaglia2013simulation,smith2018different}, and that they use these models to learn how to use new objects to accomplish novel goals in just a handful of interactions \cite{allen2020rapid}. 
Thus, we suggest that any agent that can perform flexible physical problem solving should have both prior knowledge of the dynamics of the world, as well as a way to augment those dynamics in a way that supports their interactions with the scene.
Note that we do not suggest that this specific approach corresponds to the way that humans learn or reason about physics, but instead that we believe augmented simulation is key to human sample efficiency, and therefore should be important for robotic sample efficiency as well.
Fig.~\ref{fig:learningapproach} provides an idea of the proposed approach.
}
Our testbed for this problem is a circular maze environment (CME; see Fig.~\ref{fig:learningapproach}), in which the goal is to tip and tilt the maze so as to move a marble from an outer ring into an inner circle. This is an interesting domain for studying real-time control because it is intuitively easy to pick up for people --- even children play with similar toys without prior experience with these mazes --- and yet is a complex learning domain for artificial agents due to its constrained geometry, underactuated control, nonlinear dynamics, and long planning horizon with several discontinuities~\cite{baar2019, romeres2019}. 
Adding to this challenge, the CME is a system that is usually in motion, so planning and control must be done in real-time, or else the ball will continue to roll in possibly unintended ways.

The learning approach we present in this paper falls under the umbrella of Model-Based Reinforcement Learning (MBRL). In MBRL, a task-agnostic predictive model of the system dynamics is learned from exploration data. This model is then used to synthesize a controller which is used to perform the desired task using a suitable cost function. The model in our case is represented by a physics engine that roughly describes the CME with its physical properties. Additionally, we learn the residual between the actual system and the physics system using Gaussian process regression \cite{williams2006gaussian}. Such an augmented simulator -- a combination of a physics engine and a statistical function approximator -- allows us to efficiently learn models for physical systems while using minimal domain knowledge. 


\rebuttal{
\textbf{Contributions. } Our main contributions are as follows:
\begin{itemize}
    \item We present a novel framework where a hybrid model consisting of a full physics engine augmented with a machine learning model is used to control a complex physical system using NMPC in real time.
    \item We demonstrate that our proposed approach leads to sample-efficient learning in the CME: our agent learns to solve the maze within a couple of minutes of interaction.
\end{itemize}
}
We have released our code for the CME as it is a complex, low-dimensional system that can be used to study real-time physical control\footnote{https://www.merl.com/research/license/CME}.




    \section{Related Work}\label{sec:related_work}

Our work is motivated by the recent advances in (deep) reinforcement learning to solve complex tasks in areas such as computer games~\cite{mnih2015human} and robotics~\cite{levine2016, schulman2015trust}. While these algorithms have been very successful for solving simulated tasks, their applicability in real systems is sometimes questionable due to their relative sample inefficiency. This has motivated a lot of research in the area of transferring knowledge from a simulation environment to the real world~\cite{james2017transferring, golemo2018sim, peng2018sim, baar2019}. However, most of these techniques end up being very data intensive. Here we attempt to study complex physical puzzles using model-based agents in an attempt to learn to interact with the world in a sample-efficient manner. 

Recently the robotics community has seen a surge in interest in the use of general-purpose physics engines which can represent complex, multi-body dynamics~\cite{6386109}. These engines have been developed with the intention to allow real-time control of robotic systems while using them as an approximation of the physical world. However, these simulators still cannot model or represent the physical system accurately enough for control, and this has driven a lot of work in the area of sim-to-real transfer~\cite{tobin2017domain, ramos2019bayessim}. The goal of these methods is to train an agent in simulation and then transfer them to the real system using minimum involvement of the real system during training. However, most of these approaches use a model-free learning approach and thus tend to be sample inefficient. In contrast, we propose a method that trains a MBRL sim-to-real agent and thus achieves very good sample efficiency.


The idea of using residual models for model correction, or hybrid learning models for control of physical systems during learning in physical systems has also been studied in the past~\cite{hewing2019cautious, saveriano2017data, ajay2018augmenting, romeres2019, wu2012semi}.
However, most of these studies use prior physics information in the form of differential equations, which requires domain expertise and thus the methods also become very domain specific. While we rely on some amount of domain expertise and assumptions, using a general purpose physics engine to represent the physical system will allow for more readily generalization across a wide range of systems.

A similar CME has been solved with MBRL and deep reinforcement learning, in~\cite{romeres2019} and \cite{baar2019}, respectively. In~\cite{romeres2019}, the analytical equations of motion of the CME have been derived to learn a semi-parametric GP model~\cite{romeres2016online,SP_peters} of the system, and then combined with an optimal controller. In~\cite{baar2019}, a sim-to-real approach has been proposed, where a policy to control the marble(s) is learned on a simulator from images, and then transferred to the real CME. However, the transfer learning still requires a large amount of data from the real CME. 

While approaches that combine physical predictions and residuals have been used for control in the past~\cite{ajay2019combining}, here we demonstrate that this combination can be used as part of a model-predictive controller (MPC) of a much more complex system in real-time. An important point to note here is that the work presented in~\cite{ajay2019combining} uses MPC in a discrete action space, whereas for the current system we have to use nonlinear model-predictive control (NMPC) that requires a solution to a nonlinear, continuous control problem in real-time (which requires non-trivial, compute-expensive optimization)~\cite{diehl2009efficient}. Consequently, the present study deals with a more complicated learning and control problem that is relevant to a wide range of robotic systems.

    \section{Problem Formulation}\label{sec:prob_form}
We consider the problem of moving the marble to the center of the CME.
Our goal is to study the sim-to-real problem in a model-based setting where an agent uses a physics engine as its initial knowledge of the environment's physics.
Under these settings, we study and attempt to answer the following questions in the present paper. 
\begin{enumerate}
    \item What is needed in a model-based sim-to-real architecture for efficient learning in physical systems?
    \item How can we design a sim-to-real agent that behaves and learns in a data-efficient manner?
    \item \rebuttal{ How does the performance and learning of our agent compare against how humans learn to solve these tasks? }
\end{enumerate}
We use the CME as our test environment for the studies presented in this paper. However, our models and controller design are general-purpose and thus, we expect the proposed techniques could find generalized use in robotic systems. For the rest of the paper, we call the CME together with the tip-tilt platform the circular maze system (CMS). \rebuttal{At this point, we would like to note that we make some simplifications for the CMS to model actuation delays and tackle discontinuities for controller design as we describe in the following text.}

The goal of the learning agent is to learn an accurate model of the marble dynamics, that can be used in a controller, $\pi(\vu_k|\vx_k)$, in a model-predictive fashion which allows the CMS to choose an action $\vu_k$ given the state observation $\vx_k$ to drive a marble from an initial condition to the target state. We assume that the system is fully defined by the combination of the state $\vx_k$ and the control inputs $\vu_k$, and it evolves according to the dynamics $p(\vx_{k+1} | \vx_k, \vu_k)$ which are composed of the marble dynamics in the maze and the tip-tilt platform dynamics. 

As a simplification, we assume that the marble dynamics is independent of the radial dynamics in each of the individual rings, i.e., we quantize the radius of the marble position into the 4 rings of the maze.  We include the orientation of the tip-tilt platform as part of the state for our dynamical system, obtaining a five-dimensional state representation for the system, i.e., $\vx = (r_d, \beta, \gamma, \theta, \dot{\theta},)$. 
It can be noted that the radius $r_d$ is a discrete variable, whereas the rest of the state variables are continuous. The terms $\beta, \gamma$ represent the $X$ and $Y$-orientation of the maze platform, respectively, and $\theta, \dot{\theta} $ represent the angular position and velocity of the marble, measured with respect to a fixed frame of reference. Since $r_d$ is fixed for each ring of the CME, we remove $r_d$ from the state representation of the CMS for the rest of the paper. Thus, the state is represented by a four-dimensional vector $\vx =(\beta,\gamma,\theta, \dot{\theta})$. The angles $\beta, \gamma$ are measured using a laser sensor that is mounted on the tip-tilt platform (see Figure~\ref{fig:learningapproach}) while the state of the ball could be observed from a camera mounted above the CMS. For more details, interested readers are referred to~\cite{romeres2019}.

We assume that there is a discrete planner, which can return a sequence of gates that the marble can then follow to move to the center. Furthermore, from the human experiments we have observed that human subjects always try to bring the marble in front of the gate, and then tilt the CME to move it to the next ring. Therefore, we design a lower level controller to move the marble to the next ring when the marble is placed in front of the gate to the next ring. Thus, the task of the learned controller is to move the marble in a controlled way so that it can transition through the sequence of gates to reach the center of the CME. \rebuttal{This makes our underlying control problem tractable by avoiding discontinuities in the marble movement (as the marble moves from one ring to the next).}

Before describing our approach, we introduce additional nomenclature we will use in this paper. We represent the physics engine by $\fpe$, the residual dynamics model by $\fgp$, and the real system model by $\freal$, such that $\freal(\vx_k,\vu_k) \approx \fpe(\vx_k,\vu_k)+\fgp(\vx_k,\vu_k)$. We use MuJoCo~\cite{6386109} as the physics engine, however, we note that our approach is agnostic to the choice of physics engine. In the following sections, we describe how we design our sim-to-real agent in simulation, as well as on the real system.

        \begin{figure*}[t] 
        \vspace{0.05in}
        \centering
    	\includegraphics[width=0.90\textwidth]{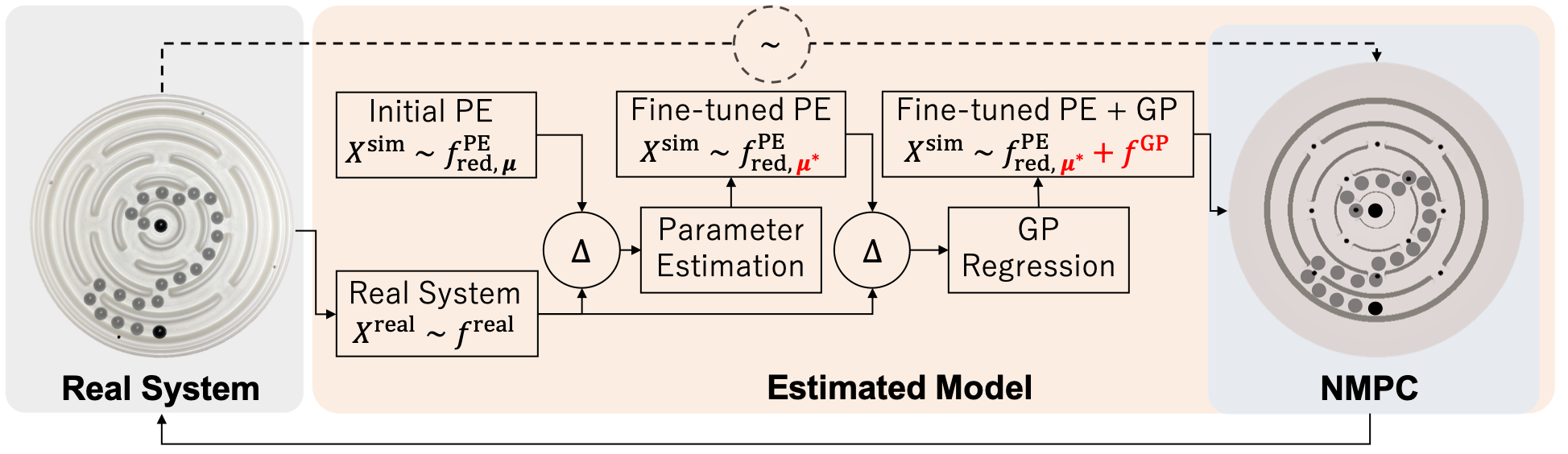}
    	\caption{
            The learning approach used in this paper to create a predictive model for the physics of the CME in the real system.
            We create a predictive model for the marble dynamics in the CME using a physics engine. We start with a MuJoCo-based physics engine (PE) with random initial parameters for dynamics, and estimate these parameters $\vmu^\ast$ from the residual error between simulated and real CME using CMA-ES. The remaining residual error between simulated and real CME is then compensated using Gaussian process (GP) regression during iterative learning. Finally, we use the augmented simulation model to control the real CME with NMPC policy. 
            %
    	}
    	\label{fig:proposed_approach}
    \end{figure*}

\section{Approach}\label{sec:approach}


Our approach for designing the learning agent is inspired by human physical reasoning: people can solve novel manipulation tasks with a handful of trials. This is mainly because we rely on already-learned notions of physics. Following a similar principle, we design an agent whose notion of physics comes from a physics engine. The proposed approach is shown as a schematic in Fig.~\ref{fig:proposed_approach}. 

We want to design a sim-to-real agent, which can bridge the gap between the simulation environment and the real world in a principled fashion. The gap between the simulated environment and the real world can be attributed to mainly two factors. First, physics engines represent an approximation of the physics of the real systems, because they are designed based on limited laws of physics, domain knowledge, and convenient approximations often made for mathematical tractability. Second, there are additional errors due to system-level problems, such as observation noise and delays, actuation noise and delays, finite computation time to update controllers based on observations, etc.

Consequently, we train our agent by first estimating the parameters of the physics engine, and then compensate for the different system-level problems as the agent tries to interact with the real system. Finally, Gaussian process Regression is used to model the residual dynamics of the real system that cannot be described by the best estimated parameters of the physics engine. In the rest of this section, we describe the details of the physics engine for the CME, and provide our approach for correcting the physics engine as well as modeling other system-level issues with the CMS.

\subsection{Physics Engine Model Description \label{subsec:physics_engine_model}}
    As described earlier, we use MuJoCo as our physics engine, $\fpe$. Note that in our model we ignore the radial movement of the marble in each ring, and describe the state only with the angular position of the marble as described in Sec.~\ref{sec:prob_form}. Consequently, we restrict the physics engine to consider only the angular dynamics of the marble in each ring, i.e., the radius of the marble position is fixed. However, in order to study the performance of the agent in simulation, we also create a full model of the CME where the marble does not have the angular state constraint. Thus, we create two different physics engine models: $\fpered$ represents the reduced physics engine available to our RL model, and $\fpefull$ uses the full internal state of the simulator. $\fpered$ differs from $\fpefull$ in two key ways.
    In the forward dynamics of the $\fpered$ model, we set the location of the marble to be in the center of each ring \rebuttal{because we cannot observe the accurate radial location of the marble in the real system, while this is tracked in $\fpefull$.}
    Additionally, because we cannot observe the spin of the ball in real experiments, we do not include it in $\fpered$, while it is included in $\fpefull$. 
    We use this $\fpefull$ model for analyzing the behavior of our agent in the preliminary studies in simulation. This serves as an analog to the real system in the simulation studies we present in the paper. We call this set of experiments \emph{sim-to-sim}. These experiments are done to determine whether the agent can successfully adapt its physics engine when initialized with an approximation of a more complicated environment.





\subsection{Model Learning}
    We consider a discrete-time system:
    \begin{equation}
        \vx_{k+1} = f(\vx_k, \vu_k) + \ve_k, \label{eq:dynamics}
    \end{equation}
    where $\vx_k \in \mathbb{R}^{4}$ denotes the state, $\vu_k \in \mathbb{R}^{2}$ the actions, and $\ve_k$ is assumed to be a zero mean white Gaussian noise with diagonal covariance, at the discrete time instant $k \in [1, ..., T]$.

    In the proposed approach, the unknown dynamics $f$ in Eq.~\ref{eq:dynamics} represents the CMS dynamics, $\freal$, and it is modeled as the sum of two components:
    \begin{equation}
        \freal(\vx_k, \vu_k) \approx \fpered(\vx_k, \vu_k) + \fgp(\vx_k, \vu_k),
    \end{equation}
    where $\fpered$ denotes the physics engine model defined in the previous section, and $\fgp$ denotes a Gaussian process model that learns the residual between real dynamics and simulator dynamics.
    We learn both the components $\fpered$ and $\fgp$ to improve model accuracy. The approach is presented as psuedo-code in Algorithm~\ref{alg:whole_learning_procedure} and described as follows.

	\begin{algorithm}[t]
		\caption{Model learning procedure}
		\label{alg:whole_learning_procedure}
		\begin{algorithmic}[1]
        \State Collect $N$ episodes in the real system using Alg.~\ref{alg:rollout}
        \State Compute simulator trajectories as $\fperedmu(\vx^\text{real}_k, \vu^\text{real}_k),$ from the real system $N$ episodes 
		\State \emph{Estimate physical parameters} using CMA-ES \label{alg:cmaes}
		\While{Model performance not converged}
		    \State Collect $N$ episodes in CMS using Alg.~\ref{alg:rollout}
		    \State Compute simulator trajectories $\vx_{k+1}^\text{sim}$ for data in $D$
		    \State \emph{Train residual GP model} \label{alg:residual}
		\EndWhile
		\end{algorithmic}
	\end{algorithm}
	\begin{algorithm}[t]
		\caption{Rollout an episode using NMPC}
 		\label{alg:rollout}
		\begin{algorithmic}[1]
		\State Initialize time index $k \leftarrow 0$
		\State Reset the real system by randomly placing the marble to outermost ring
 		\While{The marble does not reach innermost ring \textbf{and} not exceed time limit}
		    \State Set real state to simulator $\vx^\text{sim}_k \leftarrow \vx^\text{real}_k$
    		\State Compute trajectory $(X^\text{sim}, U^\text{sim})$ using NMPC
		    \State Apply initial action $\vu_k^\text{real} = \vu_0^\text{sim}$ to the real system
		    \State Store transition $D \leftarrow D \cup \{ \vx_{k}^\text{real}, \vu_{k}^\text{real}, \vx_{k+1}^\text{real}\}$
	        \State Increment time step $k \leftarrow k + 1$
		\EndWhile
		\end{algorithmic}
	\end{algorithm}

\subsubsection{Physical Parameter Estimation}
    We first estimate physical parameters of the real system.
    As measuring physical parameters directly in the real system is difficult, we estimate four friction parameters of MuJoCo by using CMA-ES~\cite{hansen2006cma}.
    More formally, we denote the physical parameters as $\vmu \in \mathbb{R}^4$, and the physics engine with the parameters as $\fperedmu$.

    As described in Algorithm~\ref{alg:whole_learning_procedure}, we first collect multiple episodes with the real system using the NMPC controller described in Sec.~\ref{subsec:NMPC}. Then, CMA-ES is used to estimate the best friction parameters $\vmu^\ast$ that minimizes the difference between the movement of the marble in the real system and in simulation as:
    \begin{equation}
        \begin{aligned}
        \vmu^\ast = \argmin_\vmu \frac{1}{\| D \|} 
        &\sum_{(\vx^\text{real}_k, \vu^\text{real}_k, \vx^\text{real}_{k+1}) \in D} \\
        &\|\vx^\text{real}_{k+1} - \fperedmu(\vx^\text{real}_k, \vu^\text{real}_k) \|^2_{W_\vmu}, 
        \label{eq:cmaes}
        \end{aligned}
    \end{equation}
    where $D$ represents the collected transitions in the real system, $W_\vmu$ is the weight matrix whose value is $1$ only related to the angular position term of the marble $\theta_{k+1}$ in the state $\vx_{k+1}$.


\subsubsection{Residual Model Learning Using Gaussian Process}
    After estimating the physical parameters, a mismatch remains between the simulator and the real system because of the modeling limitations described in the beginning of this section.
    To get a more accurate model, we train a Gaussian Process (GP) model via marginal likelihood maximization \cite{williams2006gaussian}, with a standard linear kernel, to learn the residual between the two systems by minimizing the following objective:
    \begin{equation}
        \begin{aligned}
        L^\text{GP} =& \frac{1}{\| D \|}
        \sum_{(\vx^\text{real}_k, \vu^\text{real}_k, \vx^\text{real}_{k+1}) \in D} \\ 
        &\| \left( \vx_{k+1}^\text{real} -  \fperedmuopt(\vx_{k}^\text{real}, \vu_{k}^\text{real}) \right) - \fgp(\vx_k^\text{real}, \vu_k^\text{real}) \|^2. \label{eq:gp}
        \end{aligned}
    \end{equation}
Note that after collecting the trajectories in the real system, we collect the simulator estimates of the next state $\vx_{k+1}^\text{sim}$ using the physics engine with the estimated physical parameters $\vmu^\ast$. This is done by resetting the state of the simulator to every state $\vx_{k}^\text{real}$ along the collected trajectory and applying the action $\vu_{k}^\text{real}$ to obtain the resulted next state $\vx_{k+1}^\text{sim} = \fperedmuopt(\vx_{k}^\text{real}, \vu_{k}^\text{real})$, and store the tuple $\{\vx_{k}^\text{real}, \vu_{k}^\text{real}, \vx_{k+1}^\text{sim} \}$. Thus, the GPs learn the input-output relationship: $\fgp(\vx_k^\text{real}, \vu_k^\text{real}) = \vx_{k+1}^\text{real}-\vx_{k+1}^\text{sim} $. Two independent GP models are trained, one each for the position and velocity of the marble. 
\rebuttal{We found GP models ideal for this system because of their accuracy in data prediction and data efficiency which is fundamental when working with real systems. However, other machine learning models could be adopted in different applications.}

\subsubsection{Modeling Motor Behavior}\label{sec: motor_behavior_model}
The tip-tilt platform in the CMS is actuated by hobby-grade servo motors which work in position control mode. These motors use a controller with a finite settling time which is longer than the control interval used in our experiments. This results in actuation delays for the action computed by any control algorithm, and the platform always has non-zero velocity. The physics engine, on the other hand, works in discrete time and thus the CME comes to a complete rest after completing a given action in a control interval. Consequently, there is a discrepancy between the simulation and the real system in the sense that the real system gets delayed actions. \rebuttal{Such actuation delays are common in most (robotic) control systems and thus, needs to be considered during controller design for any application}. To compensate for this problem, we learn an inverse model for motor actuation. This inverse model of the motor predicts the action to be sent to the motors for the tip-tilt platform to achieve a desired state $(\beta_{k+1}^\text{des}, \gamma_{k+1}^\text{des})$ given the current state $(\beta_k, \gamma_k)$ at instant $k$. Thus, the control signals computed by the optimization process are passed through this function that generates the commands $(u_x, u_y)$ for the servo motors. We represent this inverse motor model by $\fimm$. The motor model $\fimm$ is learned using a standard autoregressive model with external input. This is learned by collecting motor response data by exciting the CMS using sinusoidal inputs for the motors before the model learning procedure in Algorithm~\ref{alg:whole_learning_procedure}.
\subsection{Trajectory Optimization using iLQR}
    We use the iterative LQR (iLQR) as the optimization algorithm for model-based control \cite{tassa2012synthesis}. While there exist optimization solvers which can generate better optimal solutions for model-based control~\cite{betts1998survey}, we use iLQR as it provides a compute-efficient way of solving the optimization problem for designing the controller. Formally, we solve the following \emph{trajectory optimization problem} to manipulate the controls $\vu_k$ over a certain number of time steps $[T-1]$ 
    \begin{equation}
	\begin{aligned}
		\min\limits_{\vx_k,\vu_k}		&\,	\sum\limits_{k \in [T]} \ell(\vx_k,\vu_k) \\
		\text{s.t.} &\,	\vx_{k+1}=f(\vx_k,\vu_k) \\
				&\,	\vx_0  = \tilde{\vx}_0.
	\end{aligned}\label{trajopt}
\end{equation}
    For the state cost, we use a quadratic cost function for the state error measured from the target state $\vx_\text{target}$ (which in the current case is the nearest gate for the marble), as represented by the following equation:
    \begin{equation}
        \ell(\vx) =||\vx-\vx_\text{target}||_W^2, 
        \label{eq:state_cost}
    \end{equation}
    \rebuttal{where the matrix $W$ represents weights used for different states.} For the control cost, we penalize the control using a quadratic cost as well, given by the following equation:
    \begin{equation}
        \ell(\vu) = \lambda_{\vu} \| \vu \|^2.
        \label{eq:control_cost}
    \end{equation}
    
    Other smoother versions of the cost function~\cite{tassa2012synthesis} did not change the behavior of the iLQR optimization. The discrete-time dynamics $\vx_{k+1} = f(\vx_k, \vu_k)$ and the cost function are used to compute locally linear models and a quadratic cost function for the system along a trajectory.
    These linear models are then used to compute optimal control inputs and local gain matrices by iteratively solving the associated LQR problem.
    For more details of iLQR, interested readers are referred to \cite{tassa2012synthesis}. The solution to the trajectory optimization problem returns an optimal sequence of states and control inputs for the system to follow. We call this the reference trajectory for the system, denoted by  $X^\text{ref} \equiv \vx_0, \vx_1,\dots,\vx_T$, and $U^\text{ref} \equiv \vu_0, \vu_1, \dots, \vu_{T-1}$. 
    \rebuttal{The matrix $W$ used for the experiments is diagonal, $W=\textrm{diag}(4,4,1,0.4)$ and $\lambda_{\vu}=20$. These weights were tuned empirically only once at the beginning of learning.}
    
\subsection{Online Control using Nonlinear Model-Predictive Control}\label{subsec:NMPC}
While it is easy to control the movement of the marble in the simulation environment, controlling the movement of the marble in the real system is much more challenging. 
This is mainly due to complications such as static friction (which remains poorly modeled by the physics engine), or delays in actuation.
As a result, the real system requires online model-based feedback control. While re-computing an entire new trajectory upon a new observation would be the optimal strategy, due to lack of computation time in the real system, we use a trajectory-tracking MPC controller. We use an iLQR-based NMPC controller to track the trajectory obtained from the trajectory optimization module to control the system in real-time. The controller uses the least-squares tracking cost function given by the following equation:
\begin{equation}\label{eqn:mpc}
   \ell_\text{tracking}(\vx) = \|\vx_k - \vx_k^\text{ref}\|^2_Q,
\end{equation}
where $\vx_k$ is the system state at instant $k$, $\vx_k^\text{ref}$ is the reference state at instant $k$, and the matrix $Q$ is a weight matrix. \rebuttal{The matrix $Q$ and the cost coefficient for control are kept the same as during trajectory optimization}. The system trajectory is rolled out forward in time from the observed state, and the objective in Eq.~\ref{eqn:mpc} is minimized to obtain the desired control signals. 

We implement the control on both the real and the simulation environment at a control rate of $30$ Hz. As a result, there is not enough time for the optimizer to converge to the optimal feedback solution. Thus, we warm-start the optimizer with a previously computed trajectory. 
Furthermore, the derivatives during the system linearization in the backward step of iLQR and the forward rollout of the iLQR are obtained using parallel computing in order to satisfy the time constraints to compute the feedback step.

    \section{Experiments}\label{sec:experiments}
    In this section we test how our proposed approach performs on the CMS, and how it compares to human performance.


\subsection{Physical Property Estimation using CMA-ES} \label{sec:exp_cmaes}
    We first demonstrate how physical parameter estimation works in two different environments; sim-to-sim and sim-to-real settings.
    For sim-to-sim setting, we regard the full model $\fpefull$ as a real system because it contains full internal state that is difficult to observe in the real setup as described in Sec.~\ref{subsec:physics_engine_model}.
    Also, we regard the reduced model $\fpered$, which has the same state that can be observed in the real system, as a simulator.
    For $\fpered$, we start with default values given by MuJoCo, and we set smaller friction parameters to $\fpefull$ in the sim-to-sim setting, because we found the real maze board is much more slippery than what default MuJoCo's parameters would imply.
    For sim-to-real setting, we measure the difference between the real system and the reduced model~$\fpered$.

  \begin{figure}[t]
        \vspace{0.05in}
        \centering
    	\includegraphics[width=0.4\columnwidth]{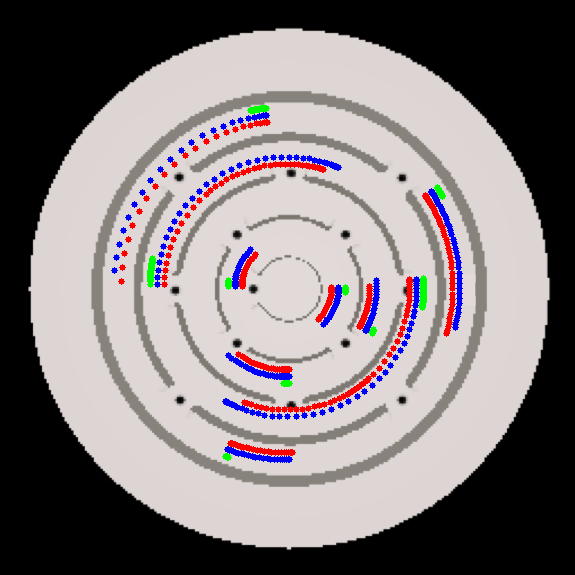}
    	\caption{Comparison of real trajectories (red), predicted trajectories (blue) using the estimated physical properties using CMA-ES, and trajectories using the default physical properties (green) in the sim-to-sim experiment. The trajectories are generated with a random policy from random initial points. 
    	}
    	\label{fig:maze_with_traj_sim}
    \end{figure}

    \rebuttal{
    To verify the performance of physical parameter estimation, we collected samples using the NMPC controller computed using current $\fpered$ models on both settings, which corresponds to line 1-3 of Algorithm~\ref{alg:whole_learning_procedure}}, and found the objective defined in \eqref{eq:cmaes} converges only $\sim 10$ transitions for each ring.
    For sim-to-sim experiment, the RMSE of ball location $\theta$ in two dynamics becomes $\approx 2e-3$ [rad] ($\approx 0.1$ [deg]), which we conclude the CMA-ES produces accurate enough parameters.
    Figure~\ref{fig:maze_with_traj_sim} shows the real trajectories obtained by $\fpefull$ (in red), simulated trajectories obtained by $\fpered$ with optimized friction parameters (in blue), and simulated trajectories before estimating friction parameters (in green).
    This qualitatively shows that the estimated friction parameters successfully bridge the gap between two different dynamics. Since tuning friction parameters for MuJoCo is not intuitive, it is evident that we can rely on CMA-ES to determine more optimal friction parameters instead.  Similarly, we find that sim-to-real experiment, the RMSE of ball position $\theta$ between the physics engine and real system decreased to $\approx 9e-3$ [rad] after CMA-ES optimization. However, we believe this error still diverges in rollout and we still suffer from static friction. We also observed that CMA-ES optimization in the sim-to-real experiments quickly finds a local minima with very few samples, and further warm starting the optimization with more data results in another set of parameters for the physics engine with similar discrepancy between the physics engine and the real system. Thus, we perform the CMA-ES parameter estimation only once in the beginning and more finetuning to GP regression.

\subsection{Control Performance on Real System} \label{subsec:sim-to-real}
We found the \textit{sim-to-sim} agent learns to perform well with just CMA-ES finetuning, and thus we skip further control results for the \textit{sim-to-sim} agent, and only present results on the real system with additional residual learning for improved performance. While CMA-ES works well in the \textit{sim-to-sim} transfer problem, if we want a robot to solve the CME, there will necessarily be differences between the internal model and real-world dynamics. We take inspiration from how people understand dynamics -- they can both capture physical properties of items in the world, and also learn the dynamics of arbitrary objects and scenes.
For this reason we augmented the CMA-ES model with machine learning data-driven models that can improve the model accuracy as more experience (data) is acquired. We opted for GP as data-driven models because of their high flexibility in describing data distribution and data efficiency~\cite{9017932}. 
    \begin{figure}[t] 
      \vspace{0.05in}
        \centering
    	\includegraphics[width=0.95\columnwidth]{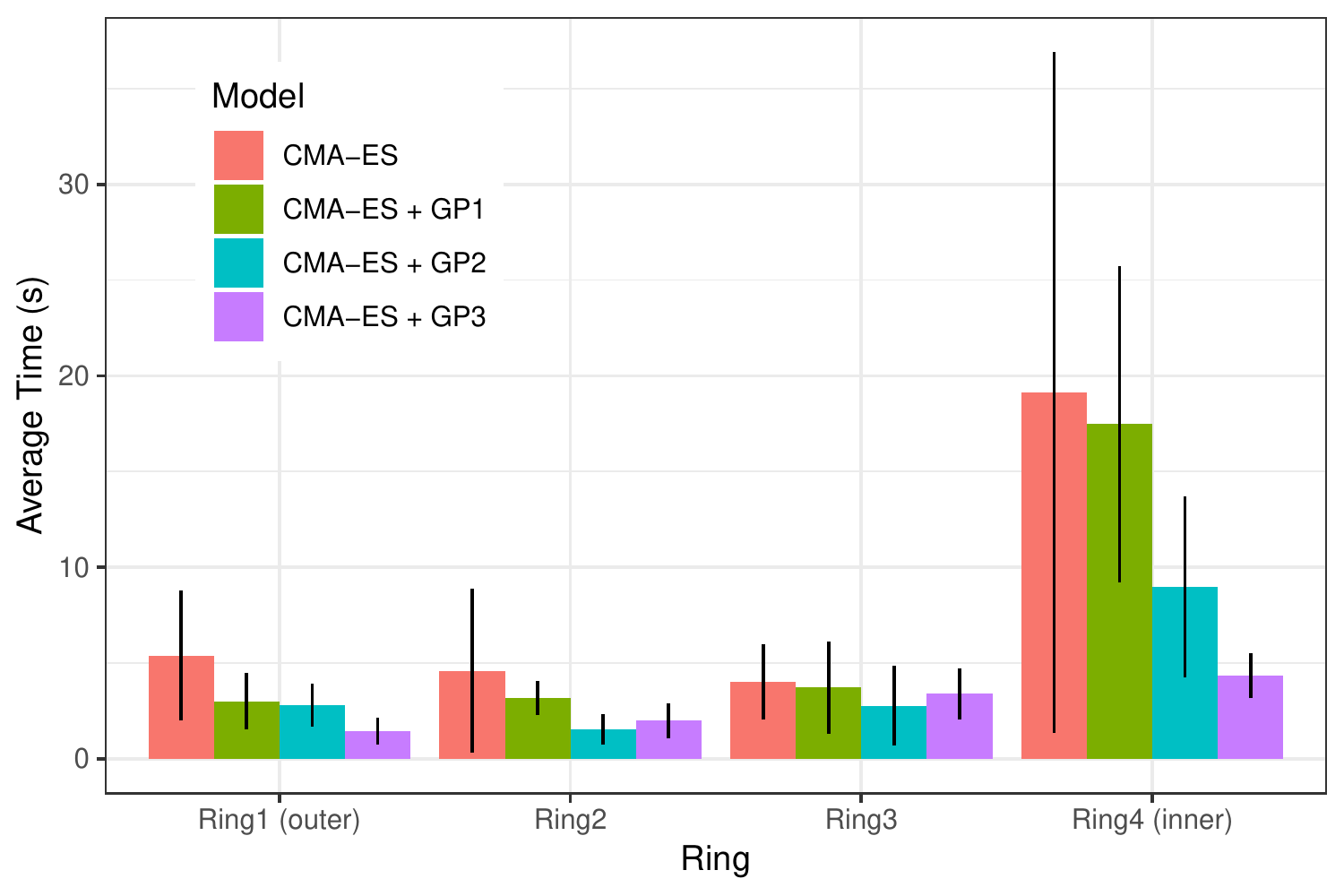}
    	\caption{Comparison of average time spent by the marble in each ring during learning \rebuttal{and the corresponding standard deviation over 10 trials}. This plot shows the improvement in the performance of the controller upon learning of the residual model. Note that the controller completely fails without CMA-ES initialization, and thus, those results are not included. }
    	\label{fig:comparison_CMAES_GP}
    \end{figure}

    
    The CMA-ES model is then iteratively improved with the GP residual model with data from $5$ rollouts in each iteration.
    In the following text, `CMA-ES' represents the CMA-ES model without any residual modeling, while `CMA-ES + GP1' represents a model that has learned a residual model from $5$ rollouts of the `CMA-ES' model. 
    Similarly, `CMA-ES + GP2' and `CMA-ES + GP3' learn the residual distribution from 10 experiments (5 with `CMA-ES' and 5 with `CMA-ES + GP1') and 15 experiments (5 each from `CMA-ES', 'CMA-ES + GP1' and `CMA-ES + GP2'), respectively. \rebuttal{The trajectory optimization and tracking uses the mean prediction from the GP models.} 
    
    Figure~\ref{fig:comparison_CMAES_GP} shows the time spent in each ring \rebuttal{averaged  over 10 different rollouts at each iteration during training.}
    As expected, models trained with a larger amount of data consistently improve the performance, i.e., spending less time in each ring. The improvement in performance can be seen especially in the outermost (Ring1; \rebuttal{$F(3,36)=3.02,~p=0.042$}) and innermost ring (Ring4; \rebuttal{$F(3,36)=4.52,~p=0.009$}).\footnote{Due to extreme heteroscedasticity in the data we use White's corrected estimators in the ANOVA \cite{white1980heteroskedasticity}.} 
    The outermost ring has the largest radius and is more prone to oscillations, which the model learns to control. Similarly, in the innermost ring, static friction causes small actions to have larger effects.
\subsection{Comparison with Human Performance}\label{sec:human_exps}

\rebuttal{
To compare our system's performance against human learning, we asked $15$ participants to perform a similar CME task. 
These participants were other members of the Mitsubishi Electric Research Laboratories who were not involved in this project and were naive to the intent of the experiment. The particpants were instructed to solve the CME five consecutive times. A 2 DoF joystick was provided to control the two servo motors of the same experimental setup on which the learning algorithm was trained. To familiarize participants with the joystick control, they were given one minute to interact with the maze---without marble. 
Because people can adapt to even unnatural joystick mappings within minutes \cite{bock2001conditions}, we assumed that this familiarization would provide a reasonable control mapping for our participants, similar to how the model pre-learned the inverse motor model $\fimm$ without learning ball dynamics.
Since we found no reliable evidence of improvement throughout the trials (see below), we believe that any further motor control learning beyond this period was at most marginal. Three participants had prior experience solving the CME in the "convential" way by holding it with both hands.
}

Afterwards, the ball was placed at a random point in the outermost ring, and participants were asked to guide the ball to the center of the maze.
They were asked to solve the CME five times, and we recorded how long they took for each solution and how much time the ball spent in each ring. Two participants were excluded from analysis because they could not solve the maze five times within the $15$ minutes allotted to them.

Because people were given five maze attempts (and thus between zero and four prior chances to learn during each attempt), we compare human performance against the CMA-ES and CMA-ES+GP1 versions of our model that have comparable amounts of training.


We find that while there was a slight numerical decrease in participants' solution times over the course of the five trials, this did not reach statistical reliability ($\chi^2(1)=1.63,~p=0.2$): participants spent an average of $110$ seconds ($95\%\text{CI}:[66,153]$) to solve the maze the first time, and $79$ seconds ($95\%\text{CI}:[38,120]$) to solve the maze the last time, and only $8$ of $13$ participants solved the maze faster on the last trial as compared to their first. 
This is similar to the learning pattern found in our model, where the solution time decreased from 33s using CMA-ES to 27s using CMA-ES+GP1, which was also not statistically reliable ($t(15)=0.56,~p=0.58$).


In addition, Table~\ref{tab:human_each_ring} shows the time that people and the model kept the ball in each ring. For statistical power we have averaged over all human attempts, and across CMA-ES and CMA-ES + GP1 to equate to human learning. In debriefing interviews, participants indicated that they found that solving the innermost ring was the most difficult, as indicated by spending more time in that ring than any others (all $ps < 0.05$ by Tukey HSD pairwise comparisons). This is likely because small movements will have the largest effect on the marble's radial position, requiring precise prediction and control. Similar to people, the model also spends the most time in the inner ring (all $ps < 0.002$ by Tukey HSD pairwise comparisons), suggesting that it shares similar prediction and control challenges to people. In contrast, a fully trained standard reinforcement learning algorithm -- the soft actor-critic  (SAC)~\cite{pmlr-v80-haarnoja18b} -- learns a different type of control policy in simulation and spends the \emph{least} amount of time in the innermost ring, since the marble has the shortest distance to travel (see Supplemental Materials for more detail). 

    \begin{table}
    \vspace{0.05in}
    	\caption{Average time spent in each ring [sec].}
        \centering
        \begin{tabular}{lcc}
            \toprule
             & Human & CMA-ES + GP0/1 \\
            \midrule
            Ring 1 (outermost ring) & 22.6 & 4.18  \\
            Ring 2 & 8.0 & 3.87\\
            Ring 3 & 24.3 & 3.85 \\
            Ring 4 (innermost ring) & 41.1 & 18.29\\
            \bottomrule
        \end{tabular}
        \label{tab:human_each_ring}
    \end{table}

    \section{Conclusions and Future Work}\label{sec:conclusion}

\rebuttal{
We take inspiration from cognitive science to build an agent that can plan its actions using an augmented simulator in order to learn to control its environment in a sample-efficient manner.
We presented a learning method for navigating a marble in a complex circular maze environment.
Learning consists of initializing a physics engine, where the physics parameters are initially estimated using the real system. The error in the physics engine is then compensated using a Gaussian process regression model which is used to model the residual dynamics.
These models are used to control the marble in the maze environment using iLQR in a feedback MPC fashion.
We showed that the proposed method can learn to solve the task of driving the marble to the center of the maze within a few minutes of interacting with the system, in contrast to traditional reinforcement systems that are data-hungry in simulation and cannot learn a good policy on a real robot.
}
    

\rebuttal{To implement our approach on the CMS, we made some simplifications that are only applicable to the CMS, e.g., that the problem can be segmented into moving through the gates in the rings. While this does limit the generality of the specific model used, most physical systems require some degree of domain knowledge to design an efficient and reliable control system. Nonetheless, we believe our approach is a step towards learning general-purpose, data-efficient controllers for complex robotic systems.} One of the benefits of our approach is its flexibility: because it learns based off of a general-purpose physics engine, this approach should generalize well to other real-time physical control tasks. Furthermore, the separation of the dynamics and control policy should facilitate transfer learning. If the maze material or ball were changed (e.g., replacing it with a small die or coin), then the physical properties and residual model would need to be quickly relearned, but the control policy should be relatively similar. \rebuttal{ In future work, we plan to test the generality and transfer of this approach to different mazes and marbles. For more effective use of physics engines for these kind of problems, we would like to interface general-purpose robotics optimization software~\cite{drake} to make it more useful for general-purpose robotics application.}

    \bibliographystyle{unsrt} 
    \bibliography{references}
    \appendix
\subsection{Control Performance on Simulation} \label{app:sac}
    In order to compare the performance of our approach and a model-free RL algorithm, we train a SAC~\cite{pmlr-v80-haarnoja18b} agent with $\fpefull$ dynamics in simulation. The hyperparameters, architectures, activation function of SAC are the same as used in~\cite{pmlr-v80-haarnoja18b}.
    We also evaluate the performance of our method in sim-to-sim setting, which omits the GP part because CMA-ES quickly matches the behavior of the simulator in the sim-to-sim setting, as described in Sec.~\ref{sec:exp_cmaes}.%
    \footnote{We attempted to train SAC on the real CME, but were unable to demonstrate any learning after three days, perhaps due to complications like the continuous action space or high control frequency. However, \cite{baar2019} demonstrated sim-to-real with transfer learning could solve a somewhat different CME, suggesting a possible additional comparison for future work.}
    
    Table.~\ref{tab:sac_results} shows the average time spent in each ring for both methods.
    The SAC model solves the maze faster than the CMA-ES algorithm, but does so by speeding the ball through each ring in approximately equal time, unlike both CMA-ES and people. This is likely because the SAC agent had extensive experience to learn its control policy: it was
    trained for five million steps on the simulator, which is equivalent to approximately two days training time if done on a real system. 
    

    \begin{table}[t]
    	\caption{Average time spent each ring in simulation [sec].}
        \centering
        \begin{tabular}{lcc}
            \toprule
             & CMA-ES & SAC \\
            \midrule
            Ring 1 (outermost ring) & 1.50 & 0.78 \\
            Ring 2 & 1.00 & 0.83\\
            Ring 3 & 2.60 & 0.86 \\
            Ring 4 (innermost ring) & 7.17 & 0.73\\
            \bottomrule
        \end{tabular}
        \label{tab:sac_results}
    \end{table}

\subsection{MuJoCo Model Setting}
    As written in Sec.~\ref{subsec:physics_engine_model}, we prepare two different physics engine models: $\fpered$ and $\fpefull$.
    Table.~\ref{tab:sim-to-sim_frictions} summarizes the friction parameters $\vmu$ used for each environment. We note that these initial parameters are optimized by CMA-ES. We set the same friction parameters to all objects in the simulator: the walls and bottom that construct the circular maze, and the marble.
    We have modeled the mass and size of the marble, and geometry of the circular maze based on our measurements of the real CME used in CMS. 

    \begin{table}[t]
    	\caption{Physical parameters used in sim-to-sim experiments. The $\fpered$ uses default parameters of MuJoCo, whereas the $\fpefull$ is more slippery, because we found that the real model is actually more slippery than what default parameters would imply \cite{mujoco-web}.}
    	\centering
        \begin{tabular}{cccc}
            \toprule
            & $\fpefull$ & $\fpered$ \\
            \midrule
            Slide friction & $1e-3$ & $1$ \\
            Spin friction & $1e-6$ & $5e-3$ \\
            Roll friction & $1e-7$ & $1e-4$ \\
            Friction loss & $1e-6$ & $0$ \\
            \bottomrule
        \end{tabular}
        \label{tab:sim-to-sim_frictions}
    \end{table}

\end{document}